\documentclass[10pt,twocolumn,letterpaper,pagebackref,breaklinks,colorlinks,allcolors=wacvblue]{article}

\usepackage{wacv}              

\definecolor{colorDOCO}{HTML}{365F88}
\definecolor{colorMIMIC}{HTML}{1C7A30}
\definecolor{colorPadChest}{HTML}{2244AA}
\definecolor{colorOCO}{HTML}{BF525F}

\definecolor{noisebg}{HTML}{EDF2F7}   
\definecolor{deltabg}{HTML}{E2E0EC}   

\newcommand{\nc}{\cellcolor{noisebg}}  
\newcommand{\dc}{\cellcolor{deltabg}}  

\newcommand{\Baseline}{\textsc{Baseline}}
\newcommand{\FullNoise}{\textsc{Full Noise Image}}

\usepackage{color}

\newcommand{\bb}{\color{black}}

\newcommand{\bl}{\color{blue}}

\usepackage{colortbl}
\definecolor{deltabg}{gray}{0.92}

\newcommand{\mycomment}[1]{}
\usepackage[accsupp]{axessibility}

\usepackage{orcidlink}

\usepackage{hyperref}
\definecolor{wacvblue}{rgb}{0.21,0.49,0.74}

\def\wacvPaperID{949} 
\def\confName{WACV}
\def\confYear{2027}

\title{\textsc{ShoViR}: A Benchmark for Evaluating Vision Shortcut Learning in Radiology Report Generation}
\author{
Filippo Ruffini\textsuperscript{1,2,*} \quad
Marco Salm\'e\textsuperscript{1,*} \quad
Rosa Sicilia\textsuperscript{3} \quad
Valerio Guarrasi\textsuperscript{1} \quad
Paolo Soda\textsuperscript{1,2}\\[2pt]
\textsuperscript{1}Unit of Artificial Intelligence and Computer Systems, Department of Engineering,\\
Università Campus Bio-Medico di Roma, Rome, Italy\\
\textsuperscript{2}Department of Diagnostics and Intervention, Radiation Physics, Biomedical Engineering,\\
Umeå University, Umeå, Sweden\\
\textsuperscript{3}UniCamillus-Saint Camillus International University of Health Sciences, Rome, Italy\\
{\tt\small \{filippo.ruffini, marco.salme, valerio.guarrasi, p.soda\}@unicampus.it}\\[2pt]
{\small \textsuperscript{*}These authors contributed equally to this work.}
}

\begin{document}
\maketitle
\begin{abstract}
Current evaluation protocols for Vision-Language Models (VLMs) in Radiology Report Generation (RRG) rely on report-level metrics that measure lexical overlap or aggregate clinical correctness. However, such metrics do not test whether individual diagnostic statements stem from the actual pathological evidence visible in the image. This allows models to achieve competitive scores by exploiting learned priors or spurious correlations, a failure mode we refer to as \emph{vision shortcut}. We introduce \textsc{ShoViR}, a benchmark for evaluating vision shortcut behavior in RRG. \textsc{ShoViR} extends two spatially annotated chest X-ray datasets, MIMIC-CXR and PadChest-GR, with per-box CheXpert labels, and defines image-level and disease-level occlusion experiments that contrast baseline performance on clean images against localized, region-specific perturbations. Comparing predictions across these conditions isolates two failure modes at the disease-class level: \emph{direct shortcuts}, where a finding persists after its visual evidence is removed, and \emph{contextual shortcuts}, where detection degrades once co-occurring pathologies are occluded despite the target region remaining intact. Benchmarking eight state-of-the-art VLMs, we find that shortcut behavior varies substantially across architectures and datasets. Models achieving the highest baseline report quality do not necessarily rank highest in spatial grounding, revealing that clinically fluent generation can coexist with shallow reliance on visual evidence. These findings expose a blind spot in current RRG evaluation and motivate region-aware assessment protocols.
The project page is available at \url{https://github.com/arco-group/ShoViR-Bench}.
\end{abstract}    

\section{Introduction}
\label{sec:introduction}

Chest radiography is the most frequently performed diagnostic imaging examination worldwide, generating a rising volume of studies that places substantial pressure on radiologist workloads~\cite{mcdonald2015effects}.
Vision-Language Models (VLMs)~\cite{li2025survey} have emerged as the most promising paradigm for alleviating this pressure through automated Radiology Report Generation (RRG), with recent systems demonstrating notable
improvements in report-level quality across multiple clinical
metrics~\cite{bannur2024maira2,chen2024chexagent,chaves2024llavarad}. 
Public benchmarks such as ReXrank~\cite{zhang2025rexrank} track this progress under shared evaluation protocols, yet all existing metrics operate at the report level, measuring n-gram overlap~\cite{papineni2002bleu,lin2004rouge}, clinical label agreement~\cite{smit2020chexbert}, entity-relation matching~\cite{jain2021radgraph}, or semantic scoring~\cite{ostmeier2024green}. 
None of them tests whether individual diagnostic statements are actually derived from the corresponding clinical findings in the image, rather than inferred from contextual cues or learned priors.
Consequently, a model can exploit \emph{vision shortcuts}, which are unintended decision rules that achieve high benchmark scores by leveraging easy-to-learn correlations in the training distribution (e.g., context or acquisition artifacts) instead of the radiographic evidence~\cite{geirhos2020shortcut}. \\
In the broader VLM literature, controlled benchmarks~\cite{lee2025vlind, chi2025chimera} have exposed vision shortcut behaviors in general-purpose models, revealing a systematic reliance on linguistic priors over visual evidence.
In the medical domain, related investigations have focused primarily on hallucination in Visual Question Answering (VQA)~\cite{chen2024detecting, royer2024medhalltune, gu2025medvh}, leaving the multi-diagnosis, free-text setting of RRG largely unexplored.
A key reason is that evaluation in RRG is inherently more demanding, it requires isolating individual diagnostic findings from natural-language prose and linking each to a specific image region, a capability that no existing benchmark provides.
Motivated by the lack of region-level evaluation in radiology, this work aims \emph{to rigorously determine whether current VLMs for RRG derive diagnostic statements from the clinically informative image evidence, or whether their apparent success is partially sustained by vision shortcut behaviors learned from spurious training correlations}. \\
To this end, we introduce \textsc{ShoViR}, a benchmark that evaluates whether diagnostic statements generated by VLMs genuinely depend on the visual evidence in the image. 
The benchmark leverages two chest X-ray datasets annotated with pathology-specific bounding boxes linking report findings to their corresponding anatomical regions. 
Using these annotations, we inject distribution-preserving noise into selected image regions, either the region containing the target pathology, or regions of co-occurring findings. 
By comparing model predictions across these complementary occlusion conditions, \textsc{ShoViR} isolates two distinct behaviors: \emph{direct shortcuts}, where a model reports a finding even after its visual evidence has been removed, 
and \emph{contextual shortcuts}, where a model's prediction degrades when co-occurring pathologies are occluded despite the target pathology region remaining visible.
The contributions of \textsc{ShoViR} are threefold. \textbf{First, } the curation of two structured metadata files that extend existing chest X-ray datasets with per-box pathology labels, unifying free-text reference reports and anatomically localized bounding boxes into a single resource for region-level visual assessment in RRG.
\textbf{Second,} the design of four complementary occlusion-based experiments that disentangle direct visual grounding from contextual co-occurrence effects, defining standardized metrics and reproducible procedures applicable to any RRG model.
\textbf{Third,} the benchmarking of eight VLMs spanning diverse architectures and training paradigms, analyzing when model outputs are driven by genuine visual evidence and when they instead rely on spurious statistical regularities.
Our framework further distinguishes predictions supported by direct visual signals from those influenced by contextual co-occurrence cues.

\section{Related Work}
\label{sec:related_work}
\noindent\textbf{Radiology Report Generation and Evaluation Methodologies.}
Automated RRG has evolved from early encoder-decoder systems~\cite{jing2018automatic,chen2020generating} to modern architectures built on LLM backbones, adapted to medicine through large-scale pretraining~\cite{li2024llavamed,sellergren2025medgemma} and instruction tuning~\cite{chen2024chexagent,bannur2024maira2,salme2026concept}. 
Crucially, progress in model design has been shaped by a parallel evolution in evaluation methodology.
Early systems were assessed almost exclusively with NLP overlap metrics such as BLEU~\cite{papineni2002bleu} and ROUGE~\cite{lin2004rouge}, which capture lexical overlap but not clinical correctness.
Acknowledging this limitation, the community introduced label-extraction tools such as
CheXbert~\cite{smit2020chexbert} and entity-relation parsers such as RadGraph~\cite{jain2021radgraph}, enabling evaluation of whether generated reports contain the right diagnostic assertions.
This shift, in turn, encouraged model-level innovations targeting clinical factuality, including longitudinal and EHR-aware generation~\cite{zhang2025libra,nicolson2025impact},
domain-specialised vision-language alignment~\cite{chaves2024llavarad,pellegrini2023radialog},
reinforcement from clinical feedback signals~\cite{tanida2023interactive,nicolson2024health}.
More recently, LLM-based scoring such as
GREEN~\cite{ostmeier2024green} has moved evaluation beyond discrete label matching by comparing generated and reference reports at the sentence level, yielding fine-grained clinical accuracy scores at the cost of additional variability introduced by the LLM. 
Nonetheless, all these metrics share a critical limitation: they assess the fidelity of generated text against a reference report, but remain agnostic to whether predicted findings are anchored to the anatomical regions where the pathology is visible, leaving the spatial understanding of RRG models entirely unexamined.

\noindent\textbf{Shortcut Learning in Vision-Language Models.}
Shortcut learning~\cite{geirhos2020shortcut}, which is the tendency of models
to rely on spurious correlations rather than on the intended
decision-relevant features, is well characterized for classification
tasks in both general computer vision~\cite{hermann2020origins, li2023whac} and medical imaging~\cite{degrave2021ai, jimenez2023detecting,
gichoya2022shortcuts, brown2023detecting, oakdenrayner2020hidden,
zech2018variable}.
Investigating analogous behaviors in VLMs is considerably harder: their larger capacity increases susceptibility to dataset artifacts, their richer output spaces complicate controlled failure isolation, and the entanglement of visual and linguistic reasoning makes it difficult to attribute a correct output to genuine image understanding.
In VQA, growing evidence shows that models can largely ignore images and still achieve high accuracy by exploiting language priors~\cite{agrawal2018dont, goyal2017making, lee2025vlind, chi2025chimera}.
In the medical domain, a different line of work has targeted hallucinations, the tendency of VLMs to generate plausible-sounding but factually unsupported statements~\cite{huang2025survey, liu2024survey}, proposing 
benchmarks for its detection in medical VLMs~\cite{royer2024medhalltune, gu2025medvh, chen2024detecting}, often alongside broader trustworthiness dimensions such as fairness, safety, and robustness~\cite{xia2024cares}. 
However, hallucination is only one observable failure mode. 
Shortcut learning represents an equally concerning behavior: a model that produces a correct diagnostic statement by relying on prevalence statistics or co-occurrence patterns rather than interpreting the underlying image is equally qungrounded, yet passes every existing hallucination benchmark undetected. 
This phenomenon has been investigated in the medical domain by Nguyen et al.~\cite{nguyen2025localizing}, who probe shortcut reliance in medical VQA by replacing segmented anatomical regions with pathology-free counterparts and measuring the resulting change in binary model answers.
However, their closed-ended protocol does not extend to the multi-diagnosis, free-text setting of RRG.

\begin{figure*}[!t]
  \centering

  \includegraphics[
    trim=30 30 40 100,
    clip,
    width=0.85\textwidth,
    height=0.60\textheight,
    keepaspectratio
  ]{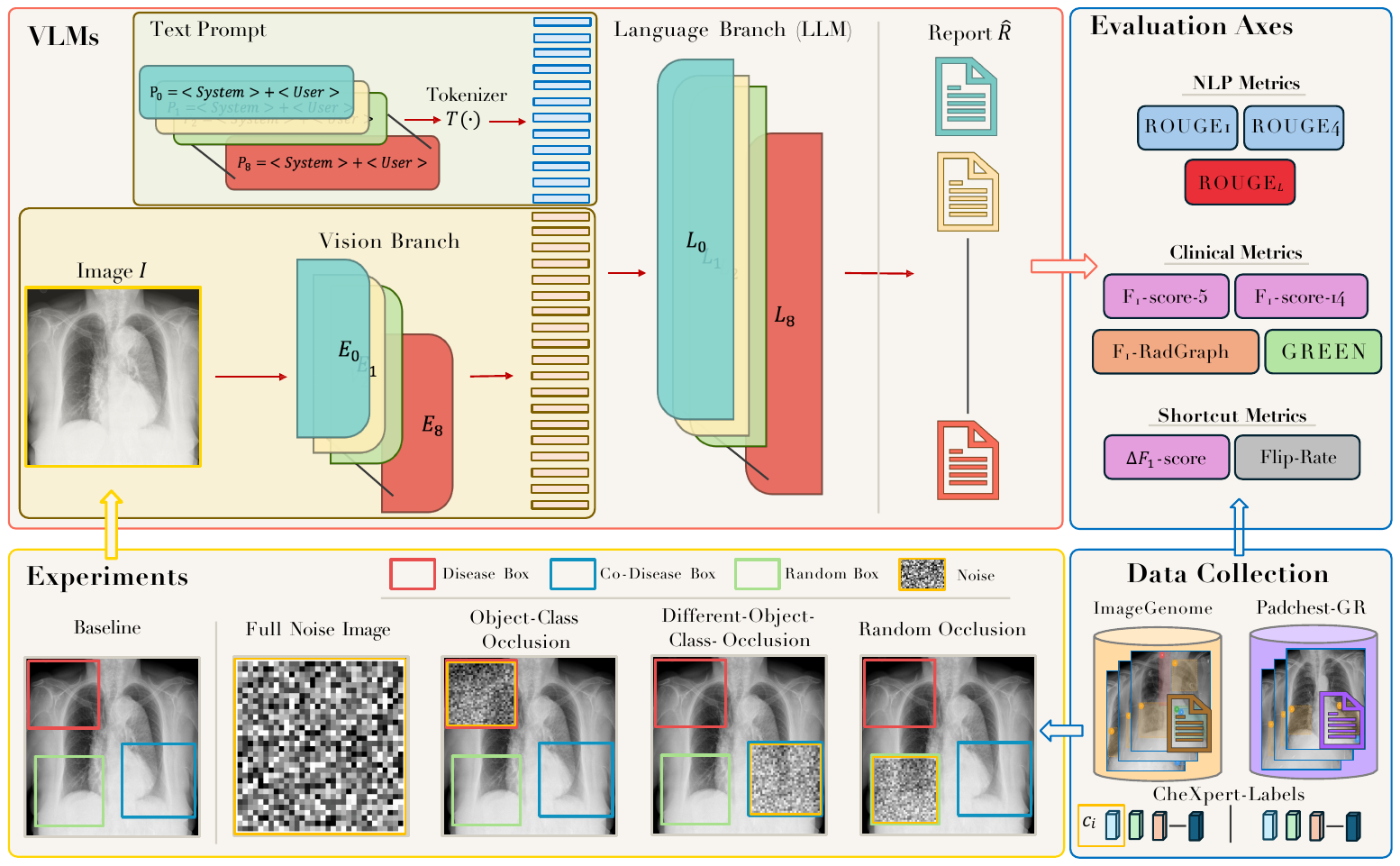}
  
  \caption{Overview of the \textsc{ShoViR} benchmark and evaluation protocol.}
  \label{fig:overview}
\end{figure*}

\section{The \textsc{ShoViR} Benchmark}

Whether VLMs ground their reports in actual disease-relevant visual evidence rather than exploiting shortcuts is essential for safe clinical deployment. 
To this end, we introduce \textsc{ShoViR} (Figure~\ref{fig:overview}), a four-stage benchmark for quantifying shortcut behaviors in VLM-based RRG designed as follows: (i)~derivation of two structured annotation sets that extend existing bounding boxes in MIMIC-CXR and PadChest-GR with per-box CheXpert labels (\S\ref{sec:dataset}); (ii)~construction of image-level and disease-level occlusion experiments that progressively remove visual evidence from the images (\S\ref{sec:experiments}); (iii)~inference framework applied to eight VLMs under a unified protocol (\S\ref{sec:models}); and (iv)~evaluation of generated reports via NLP, clinical, and shortcut metrics that measure the effect of visual evidence removal at both image
and disease-class level (\S\ref{sec:metrics}).

\bl
\mycomment{
We formalize the setting as follows.
Let~$I$ denote a chest X-ray with ground-truth diagnosis set~$\mathcal{D}_{I}$.
A VLM processes~$I$ through a vision branch~$E$ that encodes the image into a sequence of visual tokens~$\mathbf{v} = E_v(I)$, encompassing both the visual encoder and the multimodal adapter, and a language branch that tokenizes a text prompt~$P$ into text tokens~$\mathbf{t} = T(P)$.
A language model~$L$ then auto-regressively generates the output report from the concatenated token sequence:
\begin{equation}\label{eq:vlm}
  \hat{R} \;=\; L\!\bigl([\mathbf{v},\, \mathbf{t}]\bigr).
\end{equation}
Standard evaluation measures agreement between~$\hat{R}$ and a reference report at the document level.
\textsc{ShoViR} complements this by comparing~$\hat{R}$ under baseline, full-noise, and region-level occlusion conditions, exposing predictions that remain correct even when the corresponding visual evidence has been removed.
}
\bb
\subsection{Dataset Collection Process}
\label{sec:dataset}
\mycomment{
Our evaluation requires chest X-rays jointly annotated with pathology labels, radiology reports, and bounding boxes linking each finding to its anatomical region. We derived two sets from \emph{MIMIC-CXR} and \emph{PadChest-GR}, unified under the CheXpert~\cite{irvin2019chexpert} label system. For each CheXpert pathology class, we assemble a per-class image subset containing all images with at least one annotated bounding box for that class\footnote{Annotation files and the code to reproduce the data pre-processing pipeline will be available upon publication at [anonymized].}.
}
Our evaluation requires chest X-ray images annotated with three layers: radiology reports, spatially grounded bounding boxes linking each finding to its anatomical region, and per-box pathology labels.
\emph{MIMIC-CXR} and \emph{PadChest-GR} differ only in labeling conventions. Therefore, we assign CheXpert~\cite{irvin2019chexpert} labels to each box, yielding two metadata files that unify the label space for both sources. \footnote{Code to derive the annotation files will be released upon publication at [anonymized]. 
The \href{https://physionet.org/content/mimic-cxr-jpg/2.1.0/}{MIMIC-CXR}, \href{https://physionet.org/content/chest-imagenome/1.0.0/}{ChestImaGenome}, 
and \href{https://bimcv.cipf.es/bimcv-projects/padchest/}{PadChest-GR} are 
accessible online.}

\noindent\textbf{MIMIC-CXR~\cite{johnson2019mimic}} is among the most widely adopted chest radiograph corpora, pairing images with free-text reports and CheXpert labels; we build on its official test split (3{,}858 images).
To obtain spatial annotations, we incorporate ChestImaGenome~\cite{wu2021chest}, which extends a subset of MIMIC-CXR with structured bounding boxes associating clinical findings with anatomical regions. 
The ImaGenome attribute vocabulary does not directly correspond to the 14 CheXpert classes, as overlapping radiological concepts may be interpreted and labeled differently even for the same image. 
To bridge this gap, we derive a rule-based mapping between the two taxonomies, whose correctness is enforced through successive filtering steps.
The specific mapping and derivation details are reported in the Supplementary (§A.1, Table~S1).
We first discard 1{,}397 images without ImaGenome annotations and 232 whose only positive label is \textit{No Finding}, which carries no associated bounding 
boxes, yielding 2{,}229 candidates with both image-level labels and region-level bounding boxes.
A cross-source consistency check was performed to compare, for every image-class pair, the original CheXpert labels against those derived from the mapped ImaGenome attributes; the resulting F1 of 0.935 confirms high agreement between the two independent annotation sources.
Finally, we apply strict true-positive filtering, retaining only images for which every positive CheXpert class is confirmed by at least one affirmed ImaGenome attribute, preventing noisy labels from undermining the occlusion analysis. 
The resulting subset comprises 2{,}160 images, each accompanied by a radiology report, multi-label annotations across the 13 localizable CheXpert classes, and region-level bounding boxes linking each finding to its anatomical location. \\
\noindent\textbf{PadChest-GR~\cite{de2025padchest}} comprises 4,555 studies, each paired with bounding boxes localizing individual radiological findings and box-level categorical labels describing the finding type. Of these, 3,008 studies (66.0\%) contain at least one annotated bounding box; the remaining 34.0\% lack spatial annotations and are excluded.
However, PadChest-GR labels do not follow the CheXpert convention, requiring a mapping step to align them with the unified label space.
As for MIMIC-CXR, we derive a rule-based mapping to align PadChest-GR findings with CheXpert classes: a static lookup assigns 69 of the 154 finding labels to CheXpert categories, covering 43\% of annotated mentions (details in Supplementary §A.2, Tables~S3–S4).
Each mapping is validated by running CheXbert~\cite{smit2020chexbert} on the sentence accompanying the bounding box and comparing its predicted label against the mapped class; cases where the two disagree are discarded, removing 35.3\% of the annotated boxes. 
This conservative filtering prioritizes annotation precision over coverage.
Classes with fewer than 40 annotated images are excluded to ensure reliable per-class evaluation.
The resulting metadata file assigns a CheXpert label to each bounding box, yielding a final set of 1,390 images and 1,997 annotated regions across 8 categories.

\noindent\textbf{Radiologist validation.} To further validate the derived annotations, we sampled 5 cases per CheXpert class from each dataset (65 from MIMIC-CXR, 45 from PadChest-GR). For each case, an expert radiologist was shown the chest radiograph with its annotated bounding box, the original finding sentence, and the full report, and was asked to assign the most appropriate CheXpert label without knowledge of the rule-based mapping.
The radiologist additionally reviewed the complete mapping Tables~S1 and S4 in the Supplementary and confirmed the clinical coherence of every assignment.
\subsection{Experiments}
\label{sec:experiments}

Our evaluation addresses two research questions: (i)~whether current VLMs for RRG base their diagnostic statements on the visual content of the image, and (ii)~whether this grounding is localized to the relevant pathological region or distributed across contextual cues.
We design two sets of experiments based on the spatial scale of intervention: Image-level and Disease-Level.

\noindent\textbf{Image-Level.}\label{sec:baselines}
In the \textsc{Baseline} setting, models receive the original, unperturbed image, establishing the reference performance against which all other conditions are compared.
In the \textsc{Full Noise Image} setting, we replace the entire image with synthetic noise to quantify how much a model's outputs depend on genuine visual content.
The noise is designed to preserve low-level image statistics while removing all anatomical structure, avoiding out-of-distribution inputs that could confound the analysis.
For each image, we estimate the mean and standard deviation from pixel values between the $1^{st}$ and $99^{th}$ intensity percentiles, sample Gaussian noise with these matched statistics.
The result retains the intensity profile of the original without meaningful spatial structure, so any performance gap with the \textsc{Baseline} is attributable to the loss of genuine visual content.
The noise formulation and an occluder-type ablation ruling out fill-pattern confounds are detailed in Supplementary §B.

\noindent\textbf{Disease-Level.}\label{sec:occlusion_experiments}
While Image-Level experiments reveal whether models use visual information globally, they cannot determine whether individual diagnoses are grounded in their corresponding anatomical regions.
Leveraging the per-class evaluation subsets defined in Section~\ref{sec:dataset}, we introduce three region-level occlusion conditions: Random Occlusion~(\textsc{RO}), Object Class Occlusion~(\textsc{OCO}), and Different Object Class Occlusion~(\textsc{DOCO}).

Consider the evaluation subset for a target class~$c \in \mathcal{D}_{I}$.
For each image~$I$ in this subset, let~$\mathcal{B}$ denote its set of annotated bounding boxes, where each box~$b \in \mathcal{B}$ is associated with a label set~$\mathcal{D}_b$ that may contain multiple disease classes.
We partition~$\mathcal{B}$ into two spatially disjoint region sets with respect to~$c$:
\begin{equation}\label{eq:region_sets}
  \begin{aligned}
    \mathcal{B}_c \;&=\; \bigl\{\, b \in \mathcal{B} \mid c \in \mathcal{D}_b \,\bigr\}, \\
    \mathcal{B}_{\bar{c}} \;&=\; \bigl\{\, b \in \mathcal{B} \mid c \notin \mathcal{D}_b \;\wedge\; \max_{b' \in \mathcal{B}_c} \operatorname{IoU}(b,\, b') < 0.15 \,\bigr\}.
  \end{aligned}
\end{equation}
where~$\mathcal{B}_c$ is the set of \emph{disease boxes}, i.e. all boxes whose label set includes~$c$, and~$\mathcal{B}_{\bar{c}}$ is the set of \emph{co-disease boxes}, i.e. all boxes labeled with a pathology other than~$c$ whose Intersection-over-Union with every disease box of~$c$ falls below~$0.15$, ensuring spatial separation while retaining a sufficient number of co-occurring pathology boxes.

All three conditions inject noise into the targeted box regions following the same matched generation described in the \textsc{Full Noise Image} setting.
A blending strength~$p \in \{0\%,\, 20\%,\, 40\%,\, 60\%,\, 80\%,\, 100\%\}$ controls per-pixel linear interpolation between original pixels and noise, where~$p{=}0$ recovers the unperturbed \textsc{Baseline}.
Lastly, a fixed random seed ensures that all models receive identical perturbed inputs across experimental settings.
\noindent\textit{Random Occlusion (\textsc{RO})} serves as a control condition: stable performance under \textsc{RO} ensures that any degradation observed under subsequent conditions reflects the removal of specific diagnostic content rather than generic visual perturbation.
Noise is inserted into bounding boxes placed at random locations overlapping with neither~$\mathcal{B}_c$ nor~$\mathcal{B}_{\bar{c}}$;
the number of boxes is sampled uniformly from~$[1, |\mathcal{B}|]$, and their dimensions from the size distribution of all bounding boxes in its dataset.

\noindent\textit{Object Class Occlusion (\textsc{OCO})} targets the direct visual evidence for class~$c$ by occluding all disease boxes~$\mathcal{B}_c$ while leaving the rest of the image intact. 
A visually grounded model should show progressive detection loss for class~$c$ as~$p$ increases, whereas a model driven by learned priors or co-occurrence statistics may remain largely unaffected.

\noindent\textit{Different Object Class Occlusion (\textsc{DOCO})} applies the complementary intervention by occluding all co-disease boxes~$\mathcal{B}_{\bar{c}}$ while preserving~$\mathcal{B}_c$, removing surrounding pathological context without altering the target region. 
If detection of class~$c$ degrades despite its visual evidence remaining fully visible, it could implicate that the model relies on co-occurring priors or contextual learned relation with other pathologies.

\subsection{VLMs Inference}
\label{sec:models}

We benchmark eight state-of-the-art VLMs for RRG: CheXagent-2~\cite{chen2024chexagent}, CXRMate~\cite{nicolson2024health}, Libra-v1~\cite{zhang2025libra}, LLaVA-Rad~\cite{chaves2024llavarad}, MAIRA-2~\cite{bannur2024maira2}, MedGemma\footnote{MedGemma-1.5 is hereafter referred to as MedGemma for brevity.}~\cite{sellergren2025medgemma}, RaDialog~\cite{pellegrini2023radialog}, and NV-Reason-CXR~\cite{myronenko2025reasoning}; architectural details, repository links, and computational costs are reported in Supplementary §C (Tables~S5–S6, Figure~S1).
For every model, we adopt the authors' default inference configuration (text prompt, image pre-processing, and decoding strategy), so each operates under its intended settings.
A prompt sensitivity analysis validating this choice is reported in Supplementary §C.3.
To ensure a fair comparison, we give as input to all models only the frontal chest radiograph as input.

\subsection{Evaluation Axes}
\label{sec:metrics}
Our evaluation protocol comprises three complementary groups of metrics: NLP metrics that assess linguistic quality, clinical metrics that assess diagnostic correctness, and shortcut metrics that quantify performance changes under our controlled interventions.

\noindent\textbf{NLP Metrics.}
We adopt BLEU-1, BLEU-4~\cite{papineni2002bleu}, and ROUGE-L~\cite{lin2004rouge} as standard text-generation metrics. These capture surface-level $n$-gram and subsequence overlap but do not assess clinical correctness.

\noindent\textbf{Clinical Metrics.}
To assess diagnostic correctness, we employ three different metrics.
F1-CheXbert~\cite{smit2020chexbert} extracts pathology labels from generated and reference reports across the 14 CheXpert classes; we report F1\textsubscript{14} over all classes and F1\textsubscript{5} over the five most clinically prevalent findings.
For Disease-Level experiments, we compute a per-class CheXbert F1 and aggregate into a weighted-averaged score~$\mu$-F1, where each F1-score class contributes proportionally to its number of annotated images.
F1-RadGraph~\cite{jain2021radgraph} evaluates finer-grained clinical content by comparing entity-relation graphs extracted from both reports.
GREEN~\cite{ostmeier2024green} complements these with a semantically informed assessment via a fine-tuned LLM, producing scores that correlate more closely with expert radiologist judgments than rule-based alternatives.

\noindent\textbf{Shortcut Metrics.}
We quantify shortcut behavior by measuring how diagnostic correctness changes when visual evidence is selectively removed. Let~$F1(\cdot)$ denote the F1-CheXbert score under a given input condition. 
At the image level, we define the visual-reliance drop as follows:
\begin{equation}
  \Delta_{\mathrm{FN}} \;=\; F1_{(\text{\Baseline})} \;-\; F1_{(\text{\FullNoise})}
  \label{eq:delta_fullnoise}
\end{equation}
The \FullNoise\ condition provides a null-visual reference,
as it removes meaningful anatomical and pathological structure while preserving
low-level image statistics.

At the Disease-Level, we use the \textsc{RO} condition as a reference to isolate the effect of occluding diagnostically meaningful regions from the generic perturbation effect.
For a targeted condition $X\in\{\textsc{OCO},\textsc{DOCO}\}$, both computed at full occlusion ($p=100\%$), the corrected drop is
\begin{equation}
\Delta_{X}
\;=\;
\mu\text{-}F1_{(\textsc{RO})}
\;-\;
\mu\text{-}F1_{(X)}
\label{eq:delta_oco_doco}
\end{equation}
These deltas quantify the aggregate performance change induced by targeted occlusion relative to \textsc{RO}. 
A larger $\Delta_{\mathrm{OCO}}$ indicates greater sensitivity to removal of the annotated target regions, whereas a larger $\Delta_{\mathrm{DOCO}}$ indicates greater sensitivity to removal of spatially separated co-disease regions.

Since the attainable magnitude of these scores depends on the model's performance under \textsc{RO}, we complement them with a paired \emph{Flip Rate} defined at the individual-prediction level.
For a given model and class $c$, let $y_{i,c}$ be the reference label and $\hat{y}^{X}_{i,c}$ the CheXbert label extracted under condition $X$.
For $X\in\{\textsc{OCO},\textsc{DOCO}\}$, we define
\begin{equation}
\mathrm{FR}^{c}_{X}
=
\frac{
\left|
\left\{
i:
y_{i,c}=1,\,
\hat{y}^{\textsc{RO}}_{i,c}=1,\,
\hat{y}^{X}_{i,c}=0
\right\}
\right|
}{
\left|
\left\{
i:
y_{i,c}=1,\,
\hat{y}^{\textsc{RO}}_{i,c}=1
\right\}
\right|
}
\label{eq:flip_rate}
\end{equation}
Thus, $\mathrm{FR}$ is the fraction of correctly detected positive findings under \textsc{RO} that disappear after the targeted intervention. Unlike the raw $\Delta$, it is not algebraically bounded by the model's aggregate $\mu\text{-}F1$ under \textsc{RO}, reducing its dependence on available performance headroom.


\section{Evaluation Results}
\label{sec:results}
Results are organized into three sections. Section~\ref{sec:image-level} examines global visual reliance under full image corruption, assessing whether VLM statements depend on visual content. 
Section~\ref{sec:disease-level} narrows the intervention to targeted occlusion, testing whether such dependence is localized to pathological regions. Section~\ref{sec:shortcut-analysis} quantifies spatial and contextual dependence under the \textsc{RO} condition.

s
\begin{table*}[!t]
\centering

\label{tab:image-level}
\vspace{6pt}
\renewcommand{\arraystretch}{1}
\caption{%
\textbf{Image-level evaluation on MIMIC-CXR and PadChest-GR.}
\emph{Baseline}: performance on clean images.
\emph{Full Noise Image} (blue): performance when the entire image is replaced with noise.
\emph{$\Delta_{FN}$} (gray): absolute drop relative to baseline.
Per metric, the best baseline is in \textbf{bold}; the highest and lowest $\Delta$ are in \textbf{bold} and \underline{underlined}, respectively.
}
\resizebox{0.90\textwidth}{!}{%
\begin{tabular}{
    l
    ccccccc
    >{\columncolor{noisebg}}c >{\columncolor{noisebg}}c >{\columncolor{noisebg}}c
    >{\columncolor{noisebg}}c >{\columncolor{noisebg}}c >{\columncolor{noisebg}}c >{\columncolor{noisebg}}c
    >{\columncolor{deltabg}}c >{\columncolor{deltabg}}c
  }

\toprule
& \multicolumn{7}{c}{\textsc{Baseline}}
& \multicolumn{7}{c}{\cellcolor{noisebg}\textsc{Full Noise Image}}
& \multicolumn{2}{c}{\cellcolor{deltabg}\textsc{$\Delta_{FN}$ ($\uparrow$)}} \\
\cmidrule(lr){2-8} \cmidrule(lr){9-15} \cmidrule(lr){16-17}
Model
  & F1\textsubscript{14} & F1\textsubscript{5} & F1-RG
  & GR & B-1 & B-4 & R-L
  & \nc F1\textsubscript{14} & \nc F1\textsubscript{5} & \nc F1-RG
  & \nc GR & \nc B-1 & \nc B-4 & \nc R-L
  & \dc $\Delta_{FN}$F1\textsubscript{14} & \dc $\Delta_{FN}$F1\textsubscript{5} \\

\midrule
\multicolumn{17}{l}{\textit{MIMIC-CXR}} \\
\midrule

RaDialog
  & \textbf{0.596} & \textbf{0.619} & 0.202
  & 0.270 & 0.330 & 0.087 & 0.232
  & \nc 0.230 & \nc 0.334 & \nc 0.132
  & \nc \textbf{0.160} & \nc 0.185 & \nc 0.036 & \nc \textbf{0.177}
  & \dc 0.366 & \dc 0.286 \\

CheXagent-2
  & 0.576 & 0.615 & 0.202
  & 0.299 & 0.244 & 0.042 & 0.181
  & \nc 0.029 & \nc 0.000 & \nc 0.129
  & \nc 0.159 & \nc 0.064 & \nc 0.004 & \nc 0.116
  & \dc \textbf{0.547} & \dc \textbf{0.615} \\

LLaVA-Rad
  & 0.572 & 0.578 & 0.200
  & 0.281 & \textbf{0.333} & \textbf{0.097} & 0.242
  & \nc 0.320 & \nc 0.387 & \nc 0.128
  & \nc 0.154 & \nc 0.236 & \nc \textbf{0.045} & \nc 0.170
  & \dc \underline{0.252} & \dc \underline{0.190} \\

Libra-v1
  & 0.562 & 0.581 & 0.175
  & 0.256 & 0.260 & 0.055 & 0.207
  & \nc 0.215 & \nc 0.244 & \nc 0.086
  & \nc 0.084 & \nc 0.186 & \nc 0.031 & \nc 0.170
  & \dc 0.348 & \dc 0.337 \\

CXRMate
  & 0.577 & 0.607 & \textbf{0.253}
  & \textbf{0.315} & 0.332 & 0.093 & \textbf{0.246}
  & \nc \textbf{0.321} & \nc \textbf{0.412} & \nc \textbf{0.139}
  & \nc 0.115 & \nc \textbf{0.240} & \nc 0.024 & \nc 0.156
  & \dc 0.256 & \dc 0.194 \\

MedGemma
  & 0.555 & 0.553 & 0.113
  & 0.215 & 0.100 & 0.011 & 0.114
  & \nc 0.031 & \nc 0.000 & \nc 0.002
  & \nc 0.000 & \nc 0.091 & \nc 0.001 & \nc 0.086
  & \dc 0.524 & \dc 0.553 \\

MAIRA-2
  & 0.550 & 0.575 & 0.141
  & 0.196 & 0.215 & 0.033 & 0.166
  & \nc 0.093 & \nc 0.005 & \nc 0.080
  & \nc 0.053 & \nc 0.034 & \nc 0.001 & \nc 0.093
  & \dc 0.458 & \dc 0.569 \\

NV-Reason-CXR
  & 0.540 & 0.584 & 0.121
  & 0.165 & 0.077 & 0.012 & 0.086
  & \nc 0.230 & \nc 0.178 & \nc 0.044
  & \nc 0.019 & \nc 0.076 & \nc 0.009 & \nc 0.063
  & \dc 0.310 & \dc 0.405 \\

\midrule
\multicolumn{17}{l}{\textit{PadChest-GR}} \\
\midrule

RaDialog
  & 0.404 & 0.446 & 0.044
  & 0.102 & 0.019 & 0.002 & 0.043
  & \nc 0.164 & \nc \textbf{0.234} & \nc 0.024
  & \nc \textbf{0.040} & \nc 0.107 & \nc 0.004 & \nc \textbf{0.071}
  & \dc \underline{0.240} & \dc \underline{0.212} \\

CheXagent-2
  & 0.469 & 0.565 & 0.053
  & 0.159 & 0.109 & 0.015 & 0.100
  & \nc 0.030 & \nc 0.000 & \nc 0.020
  & \nc 0.002 & \nc 0.033 & \nc 0.000 & \nc 0.037
  & \dc \textbf{0.439} & \dc \textbf{0.565} \\

LLaVA-Rad
  & 0.441 & 0.495 & 0.054
  & 0.142 & 0.130 & 0.014 & 0.110
  & \nc \textbf{0.175} & \nc 0.201 & \nc 0.042
  & \nc 0.030 & \nc 0.100 & \nc 0.004 & \nc 0.063
  & \dc 0.266 & \dc 0.294 \\

Libra-v1
  & 0.428 & 0.495 & 0.042
  & 0.132 & 0.143 & 0.018 & 0.095
  & \nc 0.155 & \nc 0.232 & \nc 0.017
  & \nc 0.028 & \nc 0.094 & \nc 0.002 & \nc \textbf{0.071}
  & \dc 0.272 & \dc 0.263 \\

CXRMate
  & \textbf{0.519} & \textbf{0.586} & \textbf{0.145}
  & 0.212 & \textbf{0.306} & \textbf{0.096} & \textbf{0.194}
  & \nc 0.171 & \nc 0.211 & \nc \textbf{0.051}
  & \nc 0.033 & \nc 0.073 & \nc \textbf{0.006} & \nc \textbf{0.071}
  & \dc 0.348 & \dc 0.374 \\

MedGemma
  & 0.426 & 0.471 & 0.025
  & \textbf{0.287} & 0.028 & 0.004 & 0.041
  & \nc 0.030 & \nc 0.000 & \nc 0.000
  & \nc 0.000 & \nc 0.024 & \nc 0.000 & \nc 0.027
  & \dc 0.396 & \dc 0.471 \\

MAIRA-2
  & 0.474 & 0.555 & 0.122
  & 0.203 & 0.263 & 0.084 & 0.181
  & \nc 0.112 & \nc 0.007 & \nc 0.017
  & \nc 0.001 & \nc \textbf{0.162} & \nc 0.000 & \nc 0.067
  & \dc 0.362 & \dc 0.548 \\

NV-Reason-CXR
  & 0.432 & 0.447 & 0.022
  & 0.135 & 0.020 & 0.003 & 0.027
  & \nc 0.167 & \nc 0.166 & \nc 0.012
  & \nc 0.014 & \nc 0.019 & \nc 0.002 & \nc 0.029
  & \dc 0.265 & \dc 0.281 \\

\bottomrule

\end{tabular}

}%

\end{table*}
\subsection{Visual Reliance at the Image Level}
\label{sec:image-level}
\Cref{tab:image-level} reports \Baseline\ and \FullNoise\ performance for all evaluated models on both MIMIC-CXR and PadChest-GR across F1-CheXbert metrics (F1\textsubscript{14}, F1\textsubscript{5}), F1-RadGraph (F1-RG), GREEN score (GR), and NLP-based scores (BLEU-1, BLEU-4, ROUGE-L; abbreviated B-1, B-4, R-L).
We additionally report $\Delta_{\text{FN}}$ for F1\textsubscript{14} and F1\textsubscript{5}, defined as the performance drop between the two settings.
\textcolor{black}{\FullNoise\ provides a null-visual reference by removing pathological structure while preserving low-level image statistics.
All models exhibit a non-trivial gap between clean and noisy inputs, indicating sensitivity to the removal of meaningful visual content; however, the magnitude of this gap varies substantially across architectures.} CheXagent-2, MedGemma, and MAIRA-2 show the largest $\Delta_{\text{FN}}$ ($\geq 0.45$ on MIMIC-CXR; $\geq 0.36$ on PadChest-GR), with F1\textsubscript{14} dropping to 0.03-0.11 and GREEN scores approaching zero under \FullNoise.
Conversely, LLaVA-Rad shows the smallest $\Delta_{\text{FN}}$ on both datasets (0.252 and 0.266), while retaining the highest F1\textsubscript{14} under \FullNoise.
Since \FullNoise\ preserves the global intensity distribution of the original image, a small $\Delta_{\text{FN}}$ may indicate that predictions remain supported by low-level statistical cues.
CXRMate presents a noteworthy asymmetry: despite its small parameter count (0.2\,B), it achieves the strongest baseline on MIMIC-CXR in terms of F1-RG and ranks highest on PadChest-GR, yet shows one of the smallest $\Delta_{\text{FN}}$ on MIMIC-CXR (0.256). 
A plausible factor is its CXR-BERT-based reinforcement learning objective~\cite{nicolson2025impact}, which reinforces MIMIC-specific priors that sustain performance under noise; on PadChest-GR, where reporting style and population differ, these priors weaken and $\Delta_{\text{FN}}$ increases to 0.348.
NLP-based metrics (B-1, B-4, R-L) show smaller drops and less separation across models, as they reward lexical similarity and can be satisfied by generic report templates regardless of visual input. 
Notably, NV-Reason-CXR and MedGemma report lower NLP scores across both conditions, likely because their structured output format diverges from reference report style, penalizing lexical overlap independently of diagnostic accuracy.

\subsection{Disease-level occlusion analysis.}
\label{sec:disease-level}
\begin{figure*}[t]
  \centering
  \includegraphics[width=0.7\textwidth]{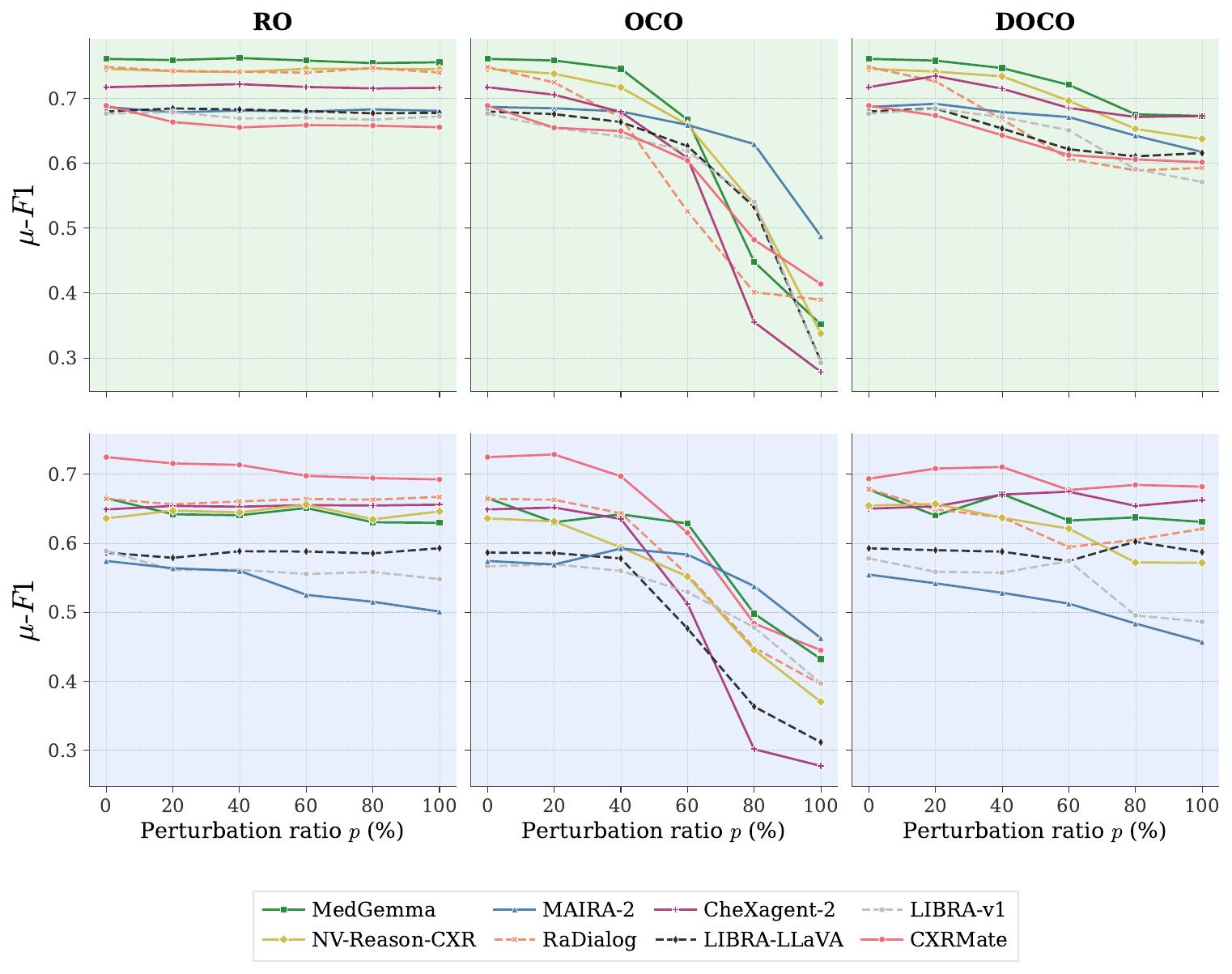}
  \caption{%
Disease-level perturbation analysis on \textcolor{colorMIMIC}{\textbf{MIMIC-CXR}} (top row) and \textcolor{colorPadChest}{\textbf{PadChest-GR}} (bottom row). 
Each curve reports the $\mu\text{-}F_1$ (weighted by class distribution) as a function of the perturbation ratio $p$ for three experimental conditions: \textsc{RO} (left panel), \textsc{OCO} (center panel), and \textsc{DOCO} (right panel). }
  \label{fig:lineplot}
\end{figure*}
While the image-level experiments reveal how strongly each model depends on global visual input, high or robust performance under noise does not necessarily imply true anatomical understanding.
The disease-level analysis determines \emph{where} in the image each model bases its predictions.
For a given disease class~$c$, we leverage its associated bounding boxes~$\mathcal{B}_c$ to selectively occlude three distinct region types: random control areas (\textsc{RO}), the target pathology regions defined by~$\mathcal{B}_c$ (\textsc{OCO}), and the regions of co-occurring pathologies~$\mathcal{B}_{\bar{c}}$ (\textsc{DOCO})\footnote{Since PadChest-GR provides multi-class bounding boxes only for a subset of images, \textsc{DOCO} is evaluated on this subset.}.
By tracking~$\mu\text{-}F1$ as a function of the perturbation ratio~$p$, we measure how each model's predictions change as specific regions are progressively corrupted.
\Cref{fig:lineplot} reports the results on both MIMIC-CXR and PadChest-GR. \\
\textsc{RO} serves as our control condition: occluded regions are placed randomly and do not overlap with disease or context-related boxes.
\textcolor{black}{Across all models, $\mu$-F1 remains largely stable as $p$ increases on both datasets, indicating limited sensitivity to the random-region perturbations used in \textsc{RO}.}
\textcolor{black}{\textsc{OCO} probes target-region sensitivity by occluding all annotated disease boxes~$\mathcal{B}_c$ associated with class~$c$.}
On both datasets, most models maintain stable $\mu$-F1 up to approximately $p{=}40\%$, then degrade sharply at higher perturbation ratios.
\textcolor{black}{This pattern indicates substantial sensitivity to the annotated target regions, although its magnitude varies across models.}
MAIRA-2 retains comparatively high $\mu$-F1 at $p{=}100\%$ (0.49 on MIMIC-CXR, 0.45 on PadChest-GR), and CXRMate remains among the strongest (0.41 on MIMIC-CXR, 0.44 on PadChest-GR), suggesting these models rely more heavily on regions outside~$\mathcal{B}_c$.
\textcolor{black}{Under \textsc{DOCO}, $\mu$-F1 also decreases for several models despite the annotated target pathology regions remaining visible.}
Since occlusion only removes co-occurring disease regions while we measure performance on the target class, this degradation indicates that predictions partly depend on contextual findings rather than solely on the target anatomy.
The contrast between stable \textsc{RO} curves and declining \textsc{DOCO} curves therefore points to a plausible shortcut behaviour: models may exploit co-occurring pathological patterns as contextual cues for predicting the target disease.

\subsection{Occlusion-based shortcut analysis}\label{sec:shortcut-analysis} 



\begin{figure*}[!t]
  \centering
  \begin{subfigure}{0.57\linewidth}
    \centering
    \includegraphics[width=\linewidth]{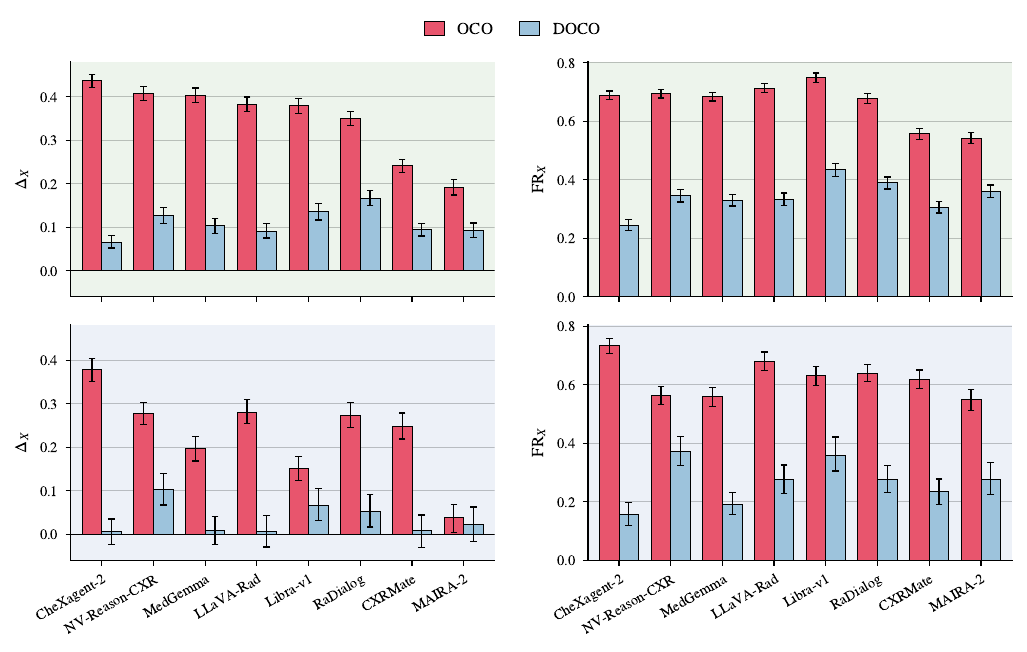}
    \caption{}
    \label{fig:occlusions_results}
  \end{subfigure}
  \hfill
  \begin{subfigure}{0.42\linewidth}
    \centering
    \includegraphics[width=\linewidth]{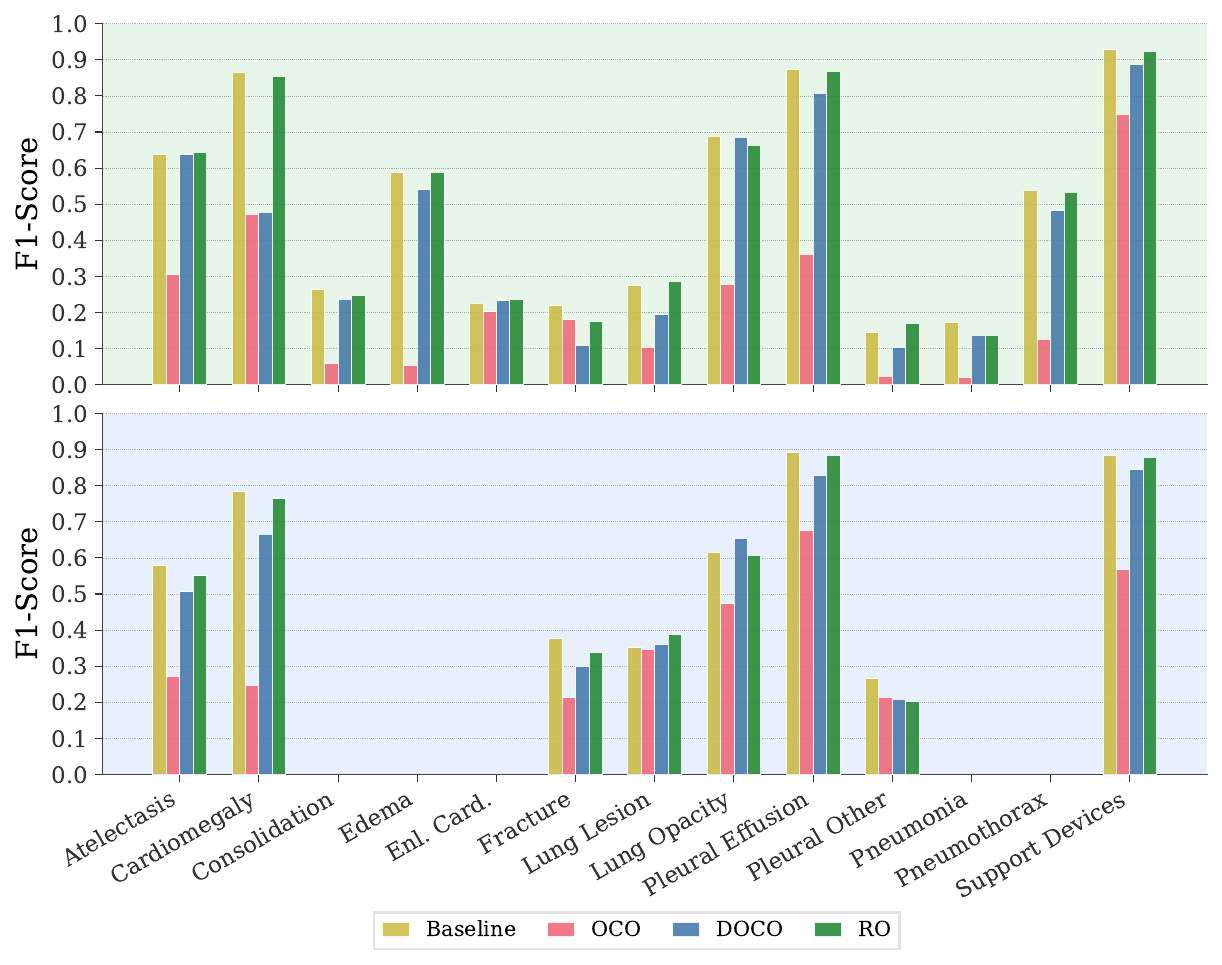}
    \caption{}
    \label{fig:disease_f1}
  \end{subfigure}
  \caption{Occlusion effects on
  \textcolor{colorMIMIC}{\textbf{MIMIC-CXR}} and \textcolor{colorPadChest}{\textbf{PadChest-GR}}.
  \textbf{(a)} Per-model delta scores $\Delta_{X}$ (left column) and Flip Rates $\mathrm{FR}_{X}$ (right column) on
  \textcolor{colorMIMIC}{\textbf{MIMIC-CXR}} (top row) and
  \textcolor{colorPadChest}{\textbf{PadChest-GR}} (bottom row), for
  $X\in\{\textcolor{colorOCO}{\textsc{OCO}},\textcolor{colorDOCO}{\textsc{DOCO}}\}$ at $p{=}100\%$.
  \textbf{(b)} Per-pathology performance averaged across all evaluated models on
  \textcolor{colorMIMIC}{\textbf{MIMIC-CXR}} (top) and
  \textcolor{colorPadChest}{\textbf{PadChest-GR}} (bottom);
  bars report $F_1$-Score for each disease category in the \Baseline\ setting and at $p{=}100\%$ for \textsc{RO}, \textsc{OCO}, and \textsc{DOCO}.}
  \label{fig:disease_occlusion}
\end{figure*}s
\Cref{fig:occlusions_results} reports the raw $\Delta_{\mathrm{F1}}$ and Flip Rate (FR) for \textsc{OCO} and \textsc{DOCO} at $p{=}100\%$ across all models and datasets.
Under \textsc{OCO}, FR ranges from 0.542 to 0.749 on MIMIC-CXR and from 0.549 to 0.734 on PadChest-GR. Every model therefore loses more than half of the findings it correctly detected under \textsc{RO}. 
On MIMIC-CXR, Libra exhibits the highest FR (0.749), followed by LLaVA-Rad (0.713). 
CheXagent-2, NV-Reason-CXR, MedGemma, and RaDialog yield similar estimates (0.677-0.694) with largely overlapping confidence intervals, whereas CXRMate and MAIRA-2 show the lowest FR values (0.557 and 0.542, respectively).
On PadChest-GR, CheXagent-2 shows the highest \textsc{OCO} FR (0.734), followed by LLaVA-Rad (0.679), while RaDialog, Libra, and CXRMate occupy an intermediate range (0.618-0.640). 
NV-Reason-CXR, MedGemma, and MAIRA-2 show lower but still substantial FR values (0.549-0.564). 
FR refines the picture given by $\Delta_{\mathrm{F1}}$. MAIRA-2 is the clearest case: its aggregate degradation falls from 0.193 on MIMIC-CXR to 0.038 on PadChest-GR, yet its FR is unchanged (0.542 vs.\ 0.549). 
Under \textsc{DOCO}, the annotated target regions remain visible, yet removing co-disease regions still induces RO$\rightarrow$0 flips in 24.5\%-43.5\% of the correctly detected findings on MIMIC-CXR.
The largest FRs are observed for Libra-v1 (0.435) and RaDialog (0.390), the lowest for CheXagent-2 (0.245), with all other models between 0.305 and 0.359.
Libra-v1 loses the largest fraction of findings correctly detected under \textsc{RO}, whereas RaDialog experiences the largest net decrease in F1 ($\Delta_{\mathrm{F1}}{=}0.168$).
On PadChest-GR, NV-Reason-CXR and Libra-v1 show the highest \textsc{DOCO} FRs (0.373 and 0.360), whereas the remaining models range from 0.158 to 0.277.
The corresponding $\Delta_{\mathrm{F1}}$ is highest for NV-Reason-CXR (0.104), followed by Libra-v1 (0.066) and RaDialog (0.054), while remaining close to zero for the other models.
Thus, even with the annotated target regions preserved, removing co-disease regions eliminates up to 37.3\% of the findings correctly detected under \textsc{RO}.
The different sensitivity profiles may partly reflect how visual evidence is incorporated during training. CheXagent-2 combines multi-task instruction tuning across 32 chest X-ray datasets and 35 tasks (e.g., abnormality detection, spatial grounding) with adaptation of the vision encoder, and exhibits high \textsc{OCO} sensitivity alongside the lowest \textsc{DOCO} sensitivity.
LLaVA-Rad lacks explicit localisation supervision but employs an image-only alignment stage to strengthen reliance on visual features, consistent with its high \textsc{OCO} FR but moderate \textsc{DOCO} response. 
The contrast with MAIRA-2 is instructive: despite receiving explicit grounding supervision, MAIRA-2 keeps its vision encoder frozen and learns localisation as part of the multimodal generation objective, yet exhibits lower \textsc{OCO} sensitivity and greater contextual dependence than CheXagent-2. 
These observations indicate that grounding supervision alone does not guarantee target-specific evidence dependence, and motivate investigating whether combining vision-encoder adaptation with diverse spatially supervised tasks can promote more selective evidence use.

To assess whether these patterns vary systematically across disease categories, \Cref{fig:disease_f1} reports the per-pathology F1 score computed independently for each disease class and averaged across all
models at full occlusion ($p{=}100\%$), comparing \textsc{RO}, \textsc{OCO}, and \textsc{DOCO} against the corresponding \Baseline\ performance.
The effect of \textsc{OCO} and \textsc{DOCO} varies considerably across both disease classes and datasets, indicating that target-region sensitivity and contextual dependence are shaped by dataset and reporting conventions rather than by inherent disease characteristics alone.
Under \textsc{OCO}, Atelectasis shows a clear performance drop on both datasets (MIMIC-CXR: 0.304 vs.\ 0.644 in \textsc{RO}; PadChest-GR: 0.271 vs.\ 0.550), whereas Fracture exhibits little separation on MIMIC-CXR but a more evident reduction on PadChest-GR, underscoring the dataset-dependent nature of the effect.
Under \textsc{DOCO}, Cardiomegaly degrades strongly on MIMIC-CXR (0.477 vs.\ 0.855 in \textsc{RO}), approaching the \textsc{OCO} drop, suggesting that for this class contextual regions carry predictive weight comparable to the target region itself.
Target-region sensitivity and contextual dependence are not intrinsic to a disease category but emerge from the interaction between co-occurrence structure and report style in the training distribution,
with prevalence playing a secondary role~\cite{irvin2019chexpert}.

\section{Conclusion}
\label{sec:conclusion}

We introduced \textsc{ShoViR}, a benchmark that evaluates visual shortcut learning in RRG through controlled occlusion experiments over two spatially annotated chest X-ray datasets. 
\textcolor{black}{Evaluating eight VLMs reveals heterogeneous patterns of target-region sensitivity and contextual dependence across models and datasets.}
Notably, high baseline report quality does not imply faithful spatial grounding: models that produce clinically fluent reports can simultaneously rely on dataset priors and co-occurrence patterns rather than on the underlying visual evidence. 
\textcolor{black}{Furthermore, target-region sensitivity and contextual dependence vary across pathologies, models, and datasets, suggesting that shortcut-related behaviors are not intrinsic to individual disease categories.}
This benchmark opens several directions for future work. Replacing noise-based occlusion with counterfactual image generation would enable stronger causal claims about visual dependence. 
Extending the analysis to probe visual token representations after the vision projector could localize whether grounding failures originate in visual encoding or language decoding. 
Finally, the contextual shortcuts exposed by our protocol motivate a finer-grained study of cross-correlated diseases and their role in driving context-dependent predictions.

{
    \small
    \bibliographystyle{ieeenat_fullname}
    \bibliography{main}
}

\end{document}